\pdfoutput=1

\documentclass[11pt]{article}

\usepackage[final]{acl}

\usepackage{times}
\usepackage{latexsym}

\usepackage[T1]{fontenc}

\usepackage[utf8]{inputenc}

\usepackage{microtype}

\usepackage{inconsolata}

\usepackage{graphicx}

\usepackage{algorithm}

\usepackage{inconsolata}
\usepackage{microtype}
\usepackage{booktabs} 
\usepackage{xspace}
\usepackage{multirow}
\usepackage{graphicx}
\usepackage{subcaption}
\usepackage{booktabs}
\usepackage[table]{xcolor}
\usepackage{amsmath}  

\usepackage{soul}

\usepackage{url}            

\usepackage{amsfonts}       
\usepackage{nicefrac}       
\usepackage{microtype}      

\usepackage{xspace}

\usepackage{bbding}

\usepackage{caption}
\usepackage{subcaption}
\usepackage{multirow}

\definecolor{lightblue}{RGB}{116,174,212}
\definecolor{lightgreen}{RGB}{211,226,183}
\definecolor{lightyellow}{RGB}{247,201,126}

\usepackage{verbatim}
\usepackage{longtable}
\usepackage{arydshln}
\usepackage{algpseudocode}
\usepackage{pifont} 
\newcommand{\ie}{\textit{i.e.,}\xspace}
\newcommand{\eg}{\textit{e.g.,}\xspace}

\newcommand{\modelname}{$D^3$\xspace}

\renewcommand{\thefootnote}{\fnsymbol{footnote}}
\newcommand{\xmark}{\ding{55}} 

\title{Position-Aware Depth Decay Decoding ($D^3$): Boosting Large Language Model Inference Efficiency}

\author{
  Siqi Fan\textsuperscript{1},
  Xuezhi Fang\textsuperscript{2}, 
  Xingrun Xing\textsuperscript{2,3},
   Peng Han\textsuperscript{1}, 
  Shuo Shang\textsuperscript{1*},
  Yequan Wang\textsuperscript{2*}\\
  $^{1}$University of Electronic Science and Technology of China, Chengdu, China\\
  $^{2}$Beijing Academy of Artificial Intelligence, Beijing, China
  \\
  $^{3}$Institute of Computing Automation, Chinese Academy of Sciences, Beijing, China \\
\texttt{\{sqfann, jedi.shang, tshwangyequan\}@gmail.com}
}

\begin{document}
\maketitle

\renewcommand{\thefootnote}{\fnsymbol{footnote}}
\footnotetext[1]{Corresponding authors.}
\renewcommand{\thefootnote}{\arabic{footnote}}

\begin{abstract}
    Due to the large number of parameters, the inference phase of Large Language Models (LLMs) is resource-intensive. Unlike traditional model compression, which needs retraining, recent dynamic computation methods show that not all components are required for inference, enabling a training-free pipeline.
    In this paper, we focus on the dynamic depth of LLM generation. A token-position aware layer skipping framework is proposed to save 1.5x times operations efficiently while maintaining performance.
    We first observed that tokens predicted later have lower perplexity and thus require less computation. Then, we propose a training-free algorithm called Position-Aware \textbf{D}epth \textbf{D}ecay \textbf{D}ecoding (\modelname), which leverages a power-law decay function, $\left\lfloor L \times (\alpha^i) \right\rfloor$, to determine the number of layers to retain when generating token $T_i$. Remarkably, without any retraining, the \modelname achieves success across a wide range of generation tasks for the first time.
    Experiments on large language models (\ie the Llama) with $7 \sim 70$ billion parameters show that \modelname can achieve an average 1.5x speedup compared with the full-inference pipeline while maintaining comparable performance with nearly no performance drop ($<1\%$) on the GSM8K and BBH benchmarks. 
    
    \end{abstract}

\section{Introduction}
\label{sec: intro}

LLMs have demonstrated impressive performance on various downstream tasks (\eg text generation, question \& answering, and sentiment analysis) using various evaluation protocols such as zero-shot, few-shot, and fine-tuning \cite{touvron2023llama}. Notably, In-context learning ability allows LLMs to adapt to tasks using input-output examples without parameter updates \cite{brown2020language,todd2023function}. However, their inference phases are very expensive due to the large number of parameters \cite{liu2023deja}. 
LLMs employ multi-layer Transformers, focusing much of the computation on decoder blocks. For LLMs like Llama, inference complexity is $LSd(d+S)$ per single inference, where $d$ is the word vector dimension, $S$ is the sequence length, and $L$ represents the number of decoder blocks \cite{DBLP:conf/sc/NarayananSCLPKV21}. This shows that computational cost scales linearly with the number of decoder blocks. 
Therefore, many works explore the possibility of achieving \textit{dynamic depth} in LLMs during inference  \cite{schuster2022confident,ma2023llm}. 

\begin{figure}[t!]
    \centering
    \includegraphics[width=\linewidth]{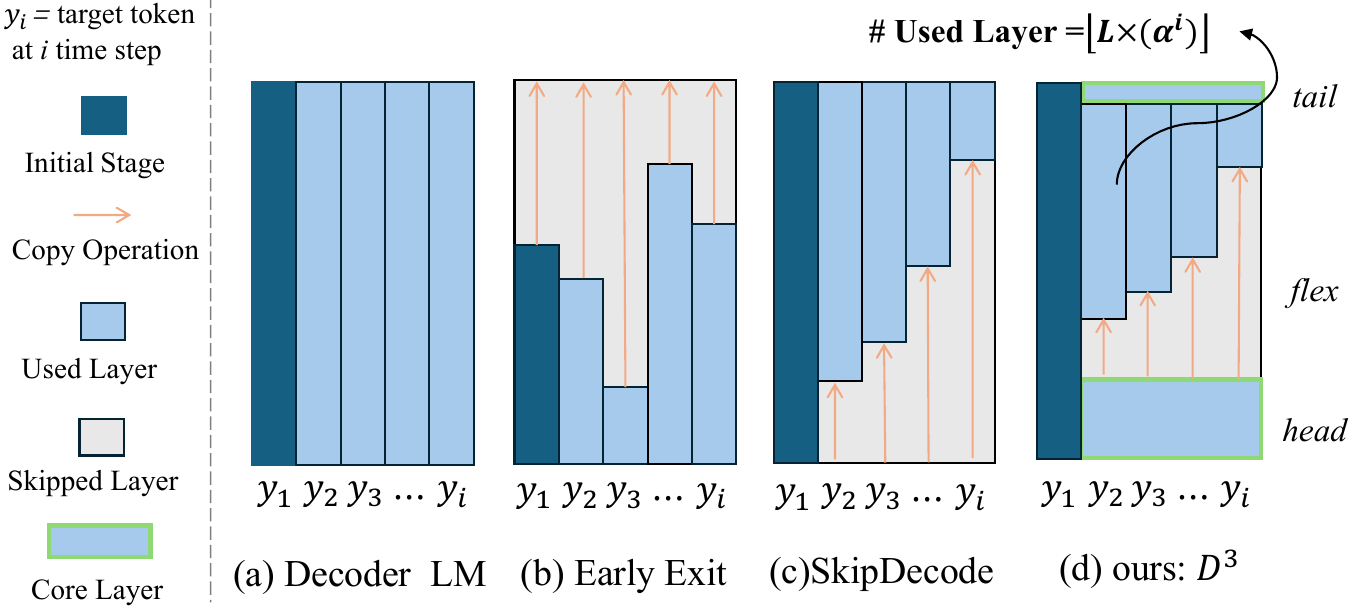}
    \caption{\modelname's generation process \textit{vs.} (a) standard implementation, (b) Early Exit, and (c) SkipDecode.}
    \label{fig:overview}
\end{figure}

Early studies in visual neural networks ~\cite{bolukbasi2017adaptive, huang2017multi} show that ``Easy'' instance activates at shallower layers while ``hard'' ones at deeper layers. \textit{Dynamic depth} was then widely used in classification tasks with encoder-only LLM like BERT \cite{li2020cascadebert,liu2020fastbert,kong2022accelerating}, mainly through early exiting and skipping layers.  The rise of generation tasks in decoder-only LLMs, prompts us to question:  ``Can we allocate different computational resources for generating tokens at different positions?"
However, extending \textit{dynamic depth} to generation tasks such as autoregressive decoding in Transformers presents challenges, as it requires computing the entire Transformer stack for each token \cite{schuster2022confident}. This introduces two key challenges:

\textbf{(\romannumeral1) Handling Missing State.}
Managing the missing hidden state and key-value cache for the current token $T_i$ becomes challenging if the previous token $T_{i-1}$ exited at a lower layer than $T_i$. The common approach \cite{corro2024skipdecode,schuster2022confident} is adding the missing state by copying. However, there are many variable factors affecting this copy operation are unclear (\eg when to add the missing state).

\textbf{(\romannumeral2) Decision on Skipping Layer Number and Direction.}
While the concept of ``easy'' instances activate at shallower layers and ``hard'' ones at deeper layers is appealing, this approach is incompatible with batching \cite{huang2017multi}. The recent SkipDecode \cite{corro2024skipdecode} addresses this by controlling the computational budget with batched exit points. However, choosing which layers to skip remains an open question. Currently, there are differing views on whether to skip head or tail layers at different timesteps. CALM \cite{schuster2022confident} uses a confidence measure to skip tail layers, while SkipDecode designs a batch exit function to skip head layers, but the reason behind this choice is unclear (Figure \ref{fig:overview}).

In this paper, we propose Position-Aware \textbf{D}epth \textbf{D}ecay \textbf{D}ecoding (\modelname) to address the above challenges. \modelname dynamically reduces the number of activated layers per token.

We find that generation performance is sensitive to early decoding steps, as shown by Llama perplexity (PPL) decrease per token. We hypothesize that \textit{``During LLM generation, tokens predicted later have lower perplexity and thus require less computation."} Previous work also \cite{schuster2022confident, corro2024skipdecode} supports this, noting that early incorrect predictions affect subsequent tokens and that average loss decreases over time. These findings guide the design of \modelname.

The core of \modelname lies in decision-making strategy for each token (\ie skip layer number and skip direction). 
Inspired by the decreasing behavior of Llama PPL per token, we design a power law decay function $ \left\lfloor L \times (\alpha^i) \right\rfloor$ to decide how many layers to keep when generating token $T_i$ at $i$ time step. Here, $L$ is the total number of layers in the model, and $\alpha$ is a hyperparameter that controls the decay rate. 
For the skip direction, previous works like \cite{DBLP:journals/corr/abs-2403-03853, yang2024laco} suggest skipping middle layers based on cosine similarity of hidden states. We further validate this by analyzing the information flow (\eg MLP, attention activations) across each block during pretraining.

Subsequently, we can perform \modelname to achieve efficient inference without additional learning. It involves only two hyperparameters: the flex layer \textit{start ID} and the \textit{decay rate ($\alpha$)}. These can be determined through grid search on a validation set on a small model and then directly transferred to larger models. Moreover, the task-specific nature of these hyperparameters allows for adaptive inference, with different parameter settings reflecting the model’s ability to adjust to specific tasks.

Experiments on well-known large language models (\ie the Llama series) with $7 \sim 70$ billion parameters show that our proposed \modelname can achieve 1.07x to 1.94x speedup compared with full-inference pipeline (\ie HuggingFace implementation) on the GSM8K and BBH benchmarks while maintaining comparable performance with nearly no performance drop ($<1\%$). More importantly, \modelname is orthogonal to other model acceleration techniques (batch processing and KV caching), offering the potential for further enhancing inference efficiency. We argue that \modelname unlocks a new paradigm for efficient inference alongside existing effective methods.

In summary, our main contributions are as follows:
\begin{itemize}
    \item A simple but effective framework (\modelname) to decide computation resource per token for LLM generation without any model retraining.
    \item A complement analysis of the token-position wise decoding that motivates the design of power law decay function. 
    \item Experiments demonstrate \modelname
     achieve average 1.5x speedup on two benchmarks while maintaining comparable performance.

\end{itemize}

\section{Related Work}
\label{sec: related work}

\paragraph{Dynamic Depth.} 
Dynamic depth involves two methods: Early Exit (EE) and Skip layer. EE first appeared in CNN/DNN networks for visual tasks \cite{bolukbasi2017adaptive, huang2017multi, teerapittayanon2016branchynet}. Subsequently, it was utilized in accelerating the inference of encoder-only architectures in BERT for classification tasks by \cite{li2020cascadebert,liu2020fastbert,li2021accelerating,kong2022accelerating}. Recently, \cite{schuster2022confident, varshney2023accelerating, corro2024skipdecode} discuss confidence-based or batch exit EE for accelerate LM inference.
Meanwhile, skip-layer dynamically omits the execution of middle layers (or modules) for any input token, facilitated by a gate function \cite{wang2018skipnet} or a binary router \cite{zeng2023learning} and layer pruning \cite{kim2024shortened,yang2024laco,song2024sleb}. 
In addition, Dola and SLED \cite{zhang2024sled,DBLP:conf/iclr/ChuangXLKGH24} improve model factuality from the perspective of dynamic depth, highlighting the potential of dynamic depth in various applications.

\paragraph{Speculative Decoding (SD).} Vanilla speculative decoding \cite{DBLP:conf/icml/LeviathanKM23,DBLP:journals/corr/abs-2302-01318} uses two models: a lightweight draft model for simple tokens and a powerful verification model for complex tokens. The draft model quickly predicts potential tokens, while the verification model checks their accuracy in parallel. 
To reduce model dependency, self-Speculative Decoding \cite{zhang2023draft,elhoushi2024layer}, using a skip-layer version of the verification model as the draft model, thus avoiding additional model loading overhead. These skipped layers are searched via Bayesian optimization, which efficiently identifies the most suitable layer configurations for draft model. Compared to existing methods, our approach eliminates secondary model loading and token verification process, with key differences summarized across four dimensions in Table \ref{tab: related diff}.

\begin{table}[t!]
    \centering
    \centering  
    
    \resizebox{\linewidth}{!}{
    \begin{tabular}{lccccc}
    \toprule
    \textbf{Method} & \textbf{Gen.} & \textbf{KV} & \textbf{Batch} & \textbf{Extra Draft} & \textbf{Task Adapt.}  \\
    \midrule
    Early-EE  & \xmark & \xmark & \xmark & \xmark & \checkmark  \\
    CLAM  & \checkmark & \xmark & \xmark & \xmark & \xmark  \\
    SkipDecode & \checkmark & \checkmark & \checkmark & \xmark & \xmark \\
    SD  & \checkmark & \checkmark & \checkmark & \checkmark & \checkmark  \\
    Self-SD & \checkmark & \checkmark & \checkmark & \xmark & \checkmark  \\
    Ours: \modelname & \checkmark & \checkmark & \checkmark & \xmark & \checkmark \\
    \bottomrule
    \end{tabular}}
    \caption{Comparison of related methods. Gen. indicates support for generation, KV and Batch for KV caching and batch processing, Extra Draft for requiring a draft model, and Task Adapt. for task adaptability.}  
    
    \label{tab: related diff}
    \end{table}

\section{Methodology}
In this section, we introduce \modelname. We begin with brief recap the two phases of text generation for decoders-only transformers for convenience (\S~\ref{sec: Preliminary}), then bring the challenges when dynamic depth comes up with generation. Last, we investigate the effects of dynamic depth on model performance (\S~\ref{sec: copy}) identify the primary reason for performance degradation, and propose strategies to mitigate them (\S~\ref{sec: core}), which guide our designed \modelname for per token-position wise decoding(\S~\ref{sec: decay}).

\subsection{Preliminary: Dissecting Efficient Inference of LLMs}
\label{sec: Preliminary}
Mainstream LLMs (\eg GPT, Llama) are rooted in the Transformer architecture \cite{DBLP:conf/nips/VaswaniSPUJGKP17}, and pretrained with a full language modeling objective with a decoder-only structure, computing loss on all tokens. 

\paragraph{Two phases behind generative tasks.}  When LLMs engage in generative tasks, they enter a next-token prediction loop until reaching an exit signal (\eg an EOS token or reaching the maximum sequence length). This process involves two phases. The first, known as the \textit{initiation phase}, entails generating the first token of the completion by processing the tokenized prompt through the network. Subsequently, the generated token is appended to the input token sequence, becoming part of the new input to generate the subsequent token. This iterative process continues until the exit signal is encountered, often referred to as the generation phase.

\paragraph{Practical accelerate methods: Batching and KV cache.} Batch processing and Key-Value (KV) cache are practical acceleration methods used during inference in LLMs. Batch processing involves simultaneously handling batches of data during inference. Since the decoder operates causally, during the generation phase, Key-Value (KV) caches prevent the recomputation of keys and values for past tokens to reduce computation costs.

\paragraph{Challenges.} 
(\romannumeral1) Instance-aware dynamic depth poses challenges for batch processing and KV cache. Confidence-based methods like CALM \cite{schuster2022confident, varshney2023accelerating} when processing batch data, instances within the same batch have to wait until the last token in a batch exits before they can stop computing. This severely restricts the practical application of such techniques \cite{corro2024skipdecode}.
(\romannumeral2) Filling missing state has error propagation. At the sequence level, if the depth at which the previous token exits is earlier than that of subsequent tokens, filling of missing layers's hidden state and KV cache for preceding tokens may cause error propagation.
(\romannumeral3) Uncertainty skip layer number and direction. CALM \cite{schuster2022confident} uses a confidence measure to skip tail layers, while SkipDecode designs a batch exit function to skip head layers. However, the rationale behind their choice is unclear.

\subsection{Error Propagation on Filling Missing State}
\label{sec: copy}

\begin{table}[t!]
	
    \centering
    
    \begin{tabular}{lccc}
    \hline Model & Layer Num.& SD & Conf   \\
    \hline 
    Llama2 7B & 32 & 21 & 0.8 \\
    Llama2 13B  & 40 & 23& 0.7\\
    Llama2 70B  & 80  &53&0.8\\
    \hline
    \end{tabular}
    \caption{LLM Saturation Statistics. ``SD" refers to the average Saturation Depth per token, and ``conf" indicates the confidence level of the Saturation Depth.}
    \label{tab: saturation} 
\end{table}

\begin{figure}[t!]
    \centering
    \includegraphics[width=\linewidth]{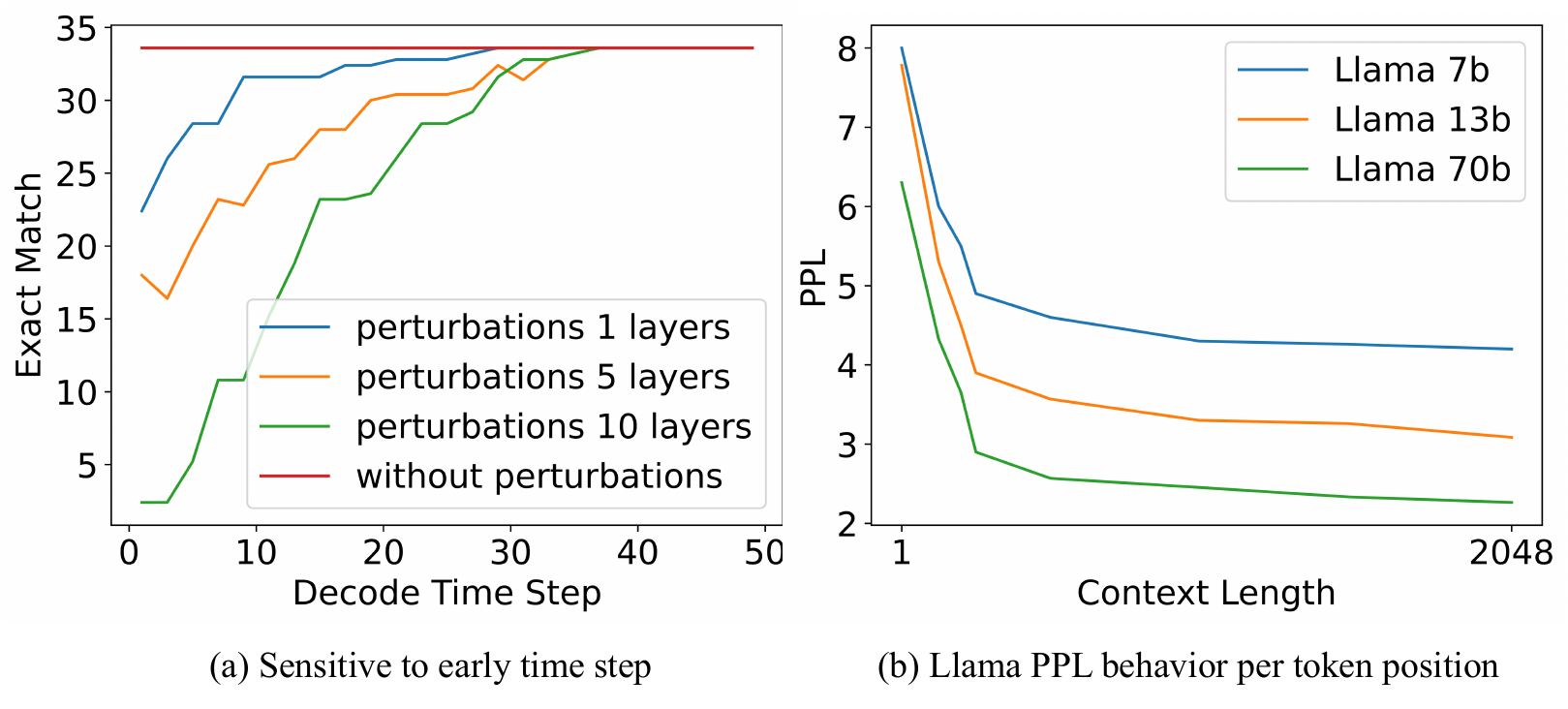}
    \caption{Error propagation and explanation from perplexity (PPL) behavior in filling missing States.}
    \label{fig:error}
\end{figure}

In autoregressive decoding, when generated token $t$, computing the input hidden state $h_{t}^i$ for layer $i$ depends on $h_{1:t-1}^{i-1}$, which is the output hidden states of the previous layer for all the tokens that have been generated so far. Therefore, if the model has early exited at some layer $j < {i - 1}$ for a token $s < t$, then $h_{s}^{i-1}$ and KV cache for $s$ is not available. 

To handle these missing hidden states and KV cache, methods like CALM \cite{elbayad2019depth, schuster2022confident} adopt the approach of copying hidden states and recomputing KV cache, while SkipDecode \cite{corro2024skipdecode} copies both hidden states and KV cache. Both methods may introduce error propagation. To this end, we first investigate the potential of layer skip in LLMs. Then we will analyze the impact of copied operation on performance, in addition to considering other factors. We use Llama2-7b \cite{touvron2023llama} and the BBH word sorting task \cite{suzgun2022challenging} for these experiments.

\begin{figure}[t!]
    \centering
    \includegraphics[width=\linewidth]{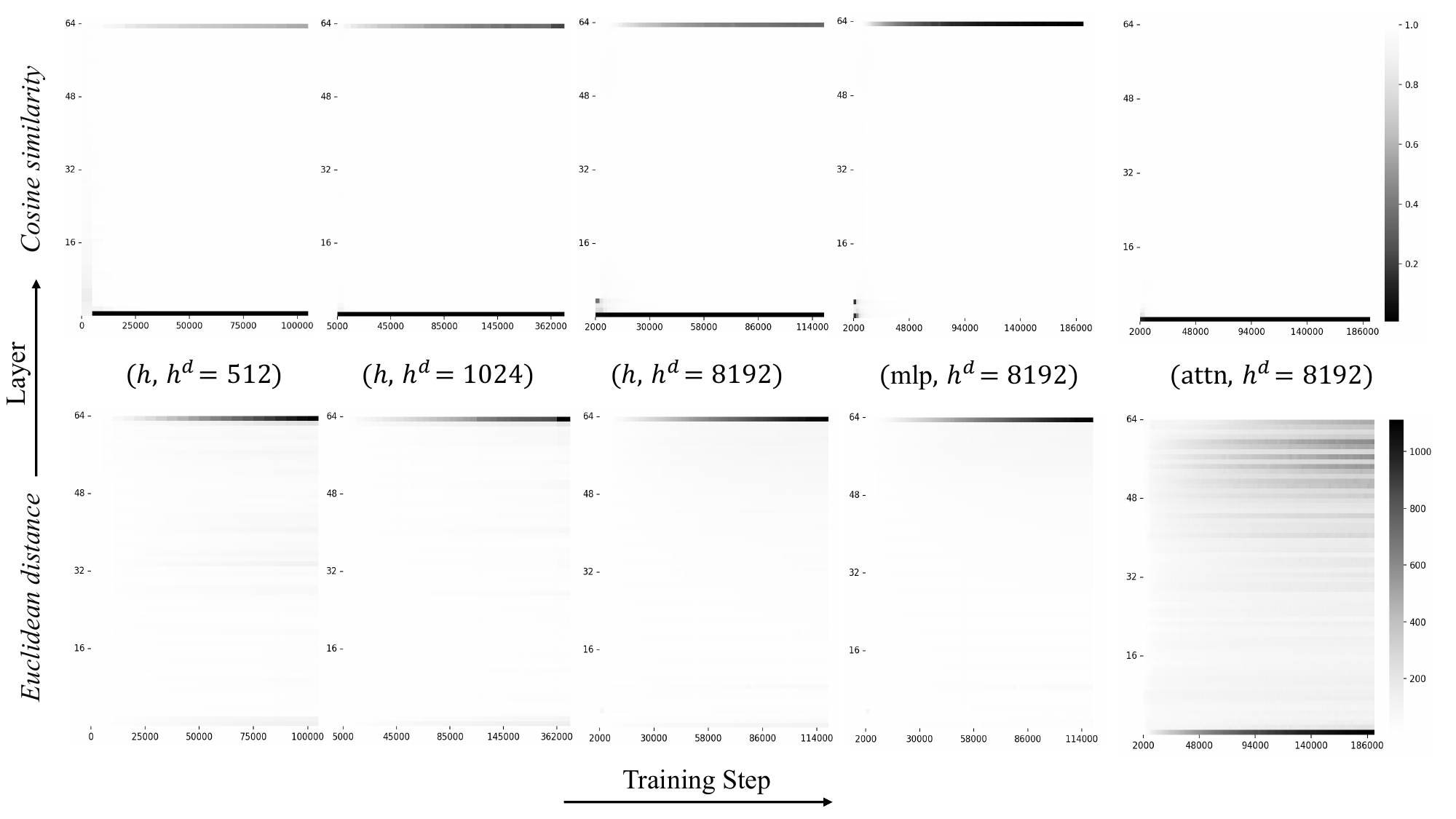}
    \caption{Visualization of input/output information flow, including features hidden state, mlp, and attention activation value, for each block during training.}
    \label{fig:cos}
\end{figure}

\paragraph{Dynamic depth potential across scaling laws.} 
\label{potential}

\begin{figure*}[t!]
    \centering
    \includegraphics[width=0.9\textwidth]{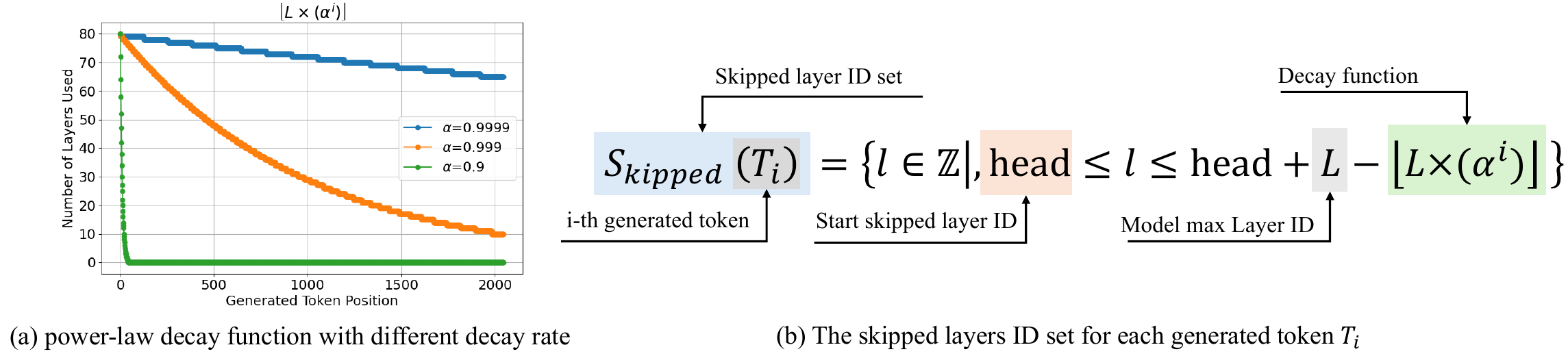}
 \caption{The layer usage for the current generated token $T_i$ follows a power-law decay function with decode time steps. }
    \label{fig:decay}
\end{figure*}

We control for the correctness of the predicted tokens to examine the potential of early exit in LLMs. That is, we first explored confidence-based oracles for decoder-only LLMs without regard to cost \cite{schuster2022confident}. Due to the exact match evaluation metric for word sorting\footnote{A given predicted string's exact match score is 1 if it is the exact same as its reference string, and is 0 otherwise.}, when generating each token $t$, we can compare each block $i$'s output hidden state $h_t^i$ with the last block hidden state $h_t^L$ after passing $lm\_head$ layer\footnote{The classification layer (\textit{lm\_head}) transforms decoder logits into a vocabulary-wide probability distribution using linear transformation and \textit{softmax}, enabling word prediction by selecting the top probability option.}.
This allows us to identify the hidden state's Saturation Depth and obtain the corresponding confidence. This process can be represented by the formula: for each generated token $t$ in each block $i, i\in [1,L)$, $\arg\max p(y_{t+1} \mid lm\_head(h_t^i)) = \arg\max p(y_{t+1} \mid lm\_head(h_t^L))$.

CALM \cite{schuster2022confident} finds that in an 8-layer T5 encoder-decoder model, exits at an average of 1.53 layers per token without performance degradation. 
Similar observation also happens in decoder-only LLMs, for the initial phase, the first token required processing through all layers, but during the generation phase, approximately $95\%$ of the tokens were correctly predicted by around $0.7$ depth of the model (Refer to Table \ref{tab: saturation}). This suggests that LLMs exhibit significant redundancy in depth, highlighting substantial potential for compute savings through layer skipping.

\paragraph{Copy operation is sensitive to early time step.}
\label{copy is sensitive}
Note that in the above Oracle experiment, missing hidden states are copied while missing KV caches are recalculated regardless of cost. When KV caches are copied, performance drops significantly (by about 20\% on the word sorting task). Although CALM \cite{schuster2022confident} addresses this by recalculating KV states, this approach is impractical for decoder-only architectures due to high computational costs \cite{corro2024skipdecode}. 
Therefore, we examine how the copy operation affects generative tasks in Figure \ref{fig:error}(a). We consider two variables: (\romannumeral1) The effect of decoder time steps (one token decoded at a time) on final performance when the same layers are copied, and (\romannumeral2) the number of layers copied. For convenience, only tail layers are copied. The results are depicted in Figure \ref{fig:error}(a), which shows that earlier perturbations lead to more significant performance degradation while copying more layers exacerbates performance decline. We attribute this outcome to perplexity (PPL) behavior per token in Figure \ref{fig:error}(b). It provides a measure of uncertainty or ``surprise" about the prediction. For a token $T_i$ with probability $p(T_i)$, it is calculated as $PPL(T_i)=\frac{1}{p(T_i)}$. At the start of the generation, limited information causes higher PPL. As more tokens are decoded, richer context lowers PPL for later tokens. Thus, we hypothesize
that \textit{``During LLM generation, tokens predicted later have
lower perplexity and thus require less computation.''}

\subsection{Core \& Flex Layer}
\label{sec: core}

Modern LLMs build coarse-grained features in their head layers and develop more detailed, fine-grained representations in deeper layers, facilitated by multi-head attention mechanisms \cite{DBLP:conf/nips/VaswaniSPUJGKP17} and residual connections \cite{he2016deep}. To investigate representation changes in each decoder block, we visualize the input-output flow during training, as shown in Figure \ref{fig:cos}. Specifically, we calculate Cosine similarity and Euclidean distances for hidden state, perception, and attention activation. For instance, given output hidden states $h_1,h_2$ from block 1 and block 2 at training step $t$,  the cosine similarity is computed as $\frac{h_1 \cdot h_2}{\|h_1\| \|h_2\|}$, and the Euclidean distance is $\sqrt{\sum_{i=1}^{d} (h_{1,i} - h_{2,i})^2}$, where $d$ is the hidden size. Other metrics, such as perception (MLP) and attention activation (attn), follow similar calculation processes. Detailed data is provided in the supplementary materials.

Results show that while middle layers exhibit minimal changes in input-output flow over time, the head and tail layers remain distinct. We propose that the head and tail layers can be called core layers, with specific roles: (\romannumeral1) head layers handle abstract, fundamental features close to the embedding layer, and (\romannumeral2) final layers align with the output near the classification layer. Middle layers, termed flex layers, are more adaptable. Dynamic depth like CALM \cite{schuster2022confident} skip tail layers, and SkipDecode \cite{corro2024skipdecode} skip head layers. We argue that the middle flex layers should be skipped first, and the optimal starting point for skipping can be determined through parameter search in smaller models.


\subsection{\modelname: Token Position Decay Strategy}
\label{sec: decay}

Previous early exit methods require a stop signal for each token, such as training early exit classifiers \cite{li2020cascadebert, liu2020fastbert} or using \textit{softmax} response \cite{schuster2022confident}. However, we believe these methods significantly increase computational load, especially since the \textit{softmax} response projects the hidden state to a large output vocabulary size. Therefore, after identifying the core layers, we propose a decay method that does not require training or \textit{softmax} response. SkipDecode \cite{corro2024skipdecode} uses linear decay given a target speedup ratio. In contrast, we design a power-law decay function $\left\lfloor L \times (\alpha^i) \right\rfloor$ based on the power-law trend of PPL behavior in Figure \ref{fig:error}(b). The decay rate is adjusted by controlling the decay coefficient $\alpha$, as illustrated in Figure \ref{fig:decay}(a). 
Given a model with $L$ layers, the skipped layers for each generated token $T_i$ are depicted in Figure \ref{fig:decay}(b).

It's worth noting that our approach is effective yet simple and easy to implement. Besides the token skipping policy, it does not necessitate any additional modifications to the transformer architecture, either during training or generation.

\section{Experiments}
\subsection{Experiment Settings}

\paragraph{Evaluation Tasks.}

We assess our method from two common text generation benchmarks from Hugging Face Open LLM Leaderboard with varying target lengths and domins: \textbf{BBH}~\cite{suzgun2022challenging}, comprising 23 challenging BIG-Bench tasks~\cite{srivastava2022beyond}, where previous language models fell short compared to human raters, necessitating multi-step reasoning and few-shot prompting without CoT; \textbf{GSM8K}~\cite{cobbe2021gsm8k}, designed for question answering on basic mathematical problems requiring multi-step reasoning. The average number of tokens in reference targets of evaluation datasets is detailed in Table \ref{tab:task}.

\paragraph{Evaluation Metrics.}
For performance evaluation, we report exact match score from the lm-harness evaluation framework~\cite{eval-harness}, this serves as the backend for Open LLM Leaderboard and is utilized by numerous open-source LLMs~\cite{biderman2023pythia,touvron2023llama}. We evaluate our approach under few-shot scenarios, using sample sizes of 3. Training set examples are added to $x_q$. For in-context learning prompts, we use a default template: $\mathrm{Q}:\left\{x_{k}\right\} \backslash \mathrm{nA}:\left\{y_{k}\right\} \backslash \mathrm{n} \backslash \mathrm{n}$, concatenating random $x_k$ and $y_k$ samples from task-specific training sets.

Following CALM~\cite{schuster2022confident}, our main efficiency metric is the average number of decoder layers used per output token, as it directly measures complexity reduction without conflating with implementation or infrastructure specific details. For reference, we convert it to average FLOPs reduction per output token \cite{elbayad2019depth,DBLP:conf/sc/NarayananSCLPKV21,corro2024skipdecode}. Taking into account the conditional checks and redundant parameter passing (\eg token position) involved in \modelname, we also compared the actual speed of \modelname in real-world scenarios compared with Hugging Face implementation, reporting wall-clock time \cite{dehghani2021efficiency}. 
\paragraph{Large Language Models.}

\begin{table}[t!]
    \centering
    \begin{minipage}[b]{0.45\linewidth}
        \centering
        \resizebox{\linewidth}{!}{
    \begin{tabular}{lccc}
    \hline Model &Params & Layer Num. &GQA  \\
    \hline 
    Llama 2 & 7B & 32  & \XSolidBrush \\
    Llama 2 & 13B  & 40 & \XSolidBrush\\
    Llama 2 & 70B & 80  &\Checkmark\\
    \hline
    \end{tabular}}
     \caption{LLMs statistics.}
     \label{tab:llm}
    \end{minipage}
    \hspace{0.05\linewidth}
    \begin{minipage}[b]{0.45\linewidth}
        \centering
        \resizebox{\linewidth}{!}{
    \begin{tabular}{lc}
    \hline Task  & output length  \\
    \hline 
    BBH & 182(2-182)\\
    GSM8K & 1230(50-1230)\\ 
    \hline
    \end{tabular}}
    \caption{Tasks statistics}
    \label{tab:task}
    \end{minipage}
\end{table}

\begin{table}[t!]
    \centering

    \resizebox{\linewidth}{!}{
    \begin{tabular}{lcc}
        \toprule
        \textbf{HP} & \textbf{Grid Search Space} & \textbf{Description} \\
        \midrule
        $start$ & [0.2, 0.3, 0.4, 0.5, 0.6, 0.7, 0.8] & Drop layer start position \\
        $\alpha$ & [0.8, 0.9, 0.999, 0.9999] & Decay rate \\
        \bottomrule
    \end{tabular}
    }
    \caption{Details of the computation layer ID set used for generating token $T_i$ at time step $i$, along with the search range and descriptions for the formula's hyperparameters.}
    \label{tab: hp search}
\end{table}

For \modelname's backbone, we choose widely recognized Llama 2 series, detailed in Table~\ref{tab:llm}. These models vary in terms of the number of parameters, ranging from 7 billion to 70 billion, and the number of layers, ranging from 32 layers to 80 layers. Compared with Llama 2 7/13B version, the 70B version employs Grouped Query Attention~\cite{ainslie2023gqa}, enhancing its inference capabilities.

\paragraph{Hyperparameter Settings}
We conducted a grid search using 10\% of the training set to determine the optimal values for $start$ and $\alpha$. The hyperparameter ranges explored are listed in Table \ref{tab: hp search}. Importantly, the optimal hyperparameters identified for smaller models were directly applied to larger models, significantly reducing the time and effort required for the search. For GSM8k, we found $\alpha=0.9999$ and $start=0.2$. For BBH, the values were $\alpha=0.99$ and $start=0.6$. Notably, these hyperparameters, optimized for the LLaMA 7B model, were efficiently transferred to the LLaMA 13B and 70B models, demonstrating a low-cost, high-efficiency approach.

\paragraph{Comparison Methods.} 
We select Early Exit \cite{schuster2022confident} (\ie CALM-DEC) and SkipDecode \cite{corro2024skipdecode} for comparison with our \modelname. The original CALM (skip deep layers) was designed for the encoder-decoder architecture of T5 and does not support batch processing due their adaptive hidden state saturation policy. To facilitate comparison, we adapted the CALM early exit concept to suit decoder-only models and enable batch processing. SkipDecode (skip shallow layers) applies linear decay within the specified upper and lower bounds of the model's executable layers, given a target speedup ratio. For a fair comparison, we ensure that SkipDecode and our method have a comparable or greater average layer number, with other parameters (\eg warm-up layer number and passing prompt inputs through all layers in the initial stage) consistent with the original paper. To ensure reproducibility, all experiments were conducted with a fixed random seed, controlling for randomness and enabling accurate comparisons.

\begin{table}[t!]
    \centering
    \setlength{\tabcolsep}{1mm}
    \resizebox{\linewidth}{!}{
        \begin{tabular}{l|ccc|ccc}
            \toprule
            \multirow{2}{*}{Methods} & \multicolumn{3}{@{}c}{{\bf BBH}} & \multicolumn{3}{@{}c}{{\bf GSM8K}} \\
            & EM & \#Avg.Layer & FLOPs r.  & EM & \#Avg.Layer & FLOPs r.  \\
            \cmidrule (r){1-1}\cmidrule (lr){2-4} \cmidrule (lr){5-7}

            \multicolumn{1}{@{}l}{{\bf \textit{Llama2 7B}}} & \multicolumn{3}{@{}l}{{ \textit{maximum length 200}}} & \multicolumn{3}{@{}l}{{ \textit{maximum length 1024}}}  \\ 
            Full Depth & 31.22 & 32.00 & -  &10.61& 32.00& - \\
            Early Exit & 31.22&19.73  &1.60x  & 10.61& 29.90 &1.07x  \\
            SkipDecode &31.65  & 24.68 & 1.28x  &  0.00 & 29.90 &1.07x \\
            \rowcolor{lightblue!30} \textbf{Ours: \modelname} &\bf{31.88} & \bf{19.17} & \bf{1.64x} &  \bf{12.05} & 29.90 & 1.07x   \\
            \hline
            \hline
            \multicolumn{1}{@{}l}{{\bf \textit{Llama2 13B}}} & \multicolumn{3}{@{}l}{{ \textit{maximum length 200}}} & \multicolumn{3}{@{}l}{{ \textit{maximum length 1024}}} \\ 
            Full Depth &37.77 &40.00  &-&22.67& 40.00& -    \\
            Early Exit &37.72& 33.50& 1.19x &    22.74& 37.51& 1.06x   \\
            SkipDecode &38.98&30.99 & 1.29x & 9.32& 37.51 &1.06x    \\
            \rowcolor{lightgreen!30}\textbf{Ours: \modelname} &\bf{39.07}&\bf{20.33} & \bf{1.94x} &  \bf{23.43} & 37.51 &1.06x  \\
            \hline
            \hline
            \multicolumn{1}{@{}l}{{\bf \textit{Llama2 70B}}} & \multicolumn{3}{@{}l}{{ \textit{maximum length 200}}} & \multicolumn{3}{@{}l}{{ \textit{maximum length 1024}}} \\ 
            Full Depth &50.90 &80.00 & -   &52.91& 80.00& -\\
            Early Exit &50.02& 50.14& 1.59x &    22.74 & 75.53& 1.05x   \\
            SkipDecode  &\bf{51.87}& 62.48& 1.27x &26.68&75.53 &1.05x  \\
            \rowcolor{lightyellow!30}\textbf{Ours: \modelname} &51.42& \bf{48.43}& \bf{1.65x} &  \bf{54.51} & 75.53 &1.05x   \\

            \bottomrule
            \end{tabular}
    }
        \caption{Main Results: Performance and efficiency across model scales.}
        \label{tab:main_results}
\end{table}

\subsection{\modelname: Performance and Efficiency Across Scaling Laws}
The main experimental results of \modelname are summarized in Tables \ref{tab:main_results} and \ref{tab: bbh em}. These experiments were conducted in few-shot settings, showcasing performance and computational efficiency compared to HuggingFace's full-depth implementation and early exit method, SkipDecode. From a perspective of performance and computational efficiency, we can draw the following experimental conclusions.

\begin{figure}[t!]
    \centering
    \includegraphics[width=\linewidth]{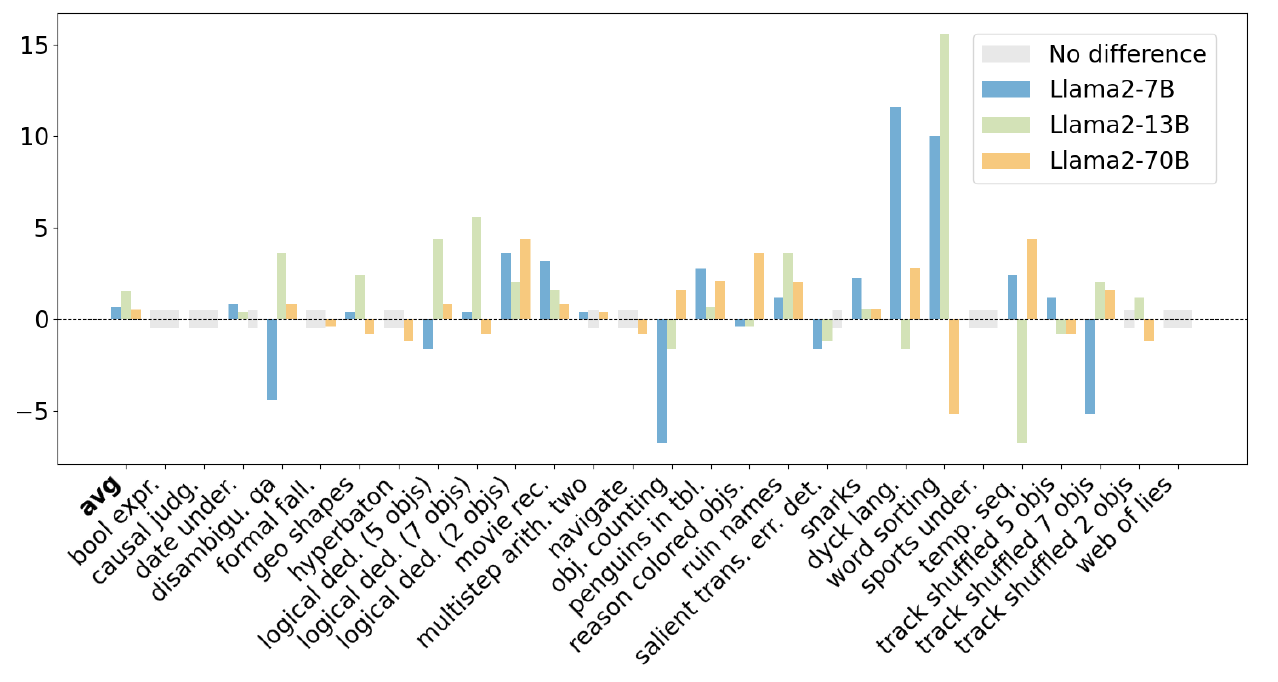}
    \caption{\modelname \textit{vs.} Full Depth performance difference (±) on BBH benchmarks}
    \label{fig:bbh bar}
\end{figure}

\paragraph{Performance is Comparable with Minimal Loss($<1\%$).}
Tables \ref{tab:main_results} and Figure \ref{fig:bbh bar} show that exact match remains within a narrow margin of $<1\%$, when compared to HuggingFace's full-depth implementation. \modelname maintains mainstream LLM capabilities and in-context learning abilities without modifying model parameters. This finding is promising, especially in light of our observation in Table \ref{tab: saturation}, where we demonstrate the feasibility of implementing early exit strategies within LLM middle layers while preserving accuracy.
For certain tasks, \modelname surpasses the last layer (Full Depth). This hints at a tendency for deep layers to potentially over-represent certain tasks, which could impede performance during LLM inference.
\paragraph{Over 1.5x Speed Up.}

\begin{table}[t!]
    \centering
    \resizebox{0.9\linewidth}{!}{
    \begin{tabular}{lccc}
    \toprule
    \textbf{Actual Time} & \textbf{Llama2 7B} & \textbf{Llama2 13B} & \textbf{Llama2 70B} \\
    \midrule
    {HF Full Depth} & 6218 & 10428 & 12893 \\
    {Ours} & 5559 & 8427 & 11294 \\
    {Speed Up} & 1.12x & 1.23x & 1.14x \\
    \bottomrule
    \end{tabular}}
    \caption{Wall-clock time(s) and actual speed up on GSM8K.}
    \label{tab: wall clock}
\end{table}

\begin{figure}[t!]
    \centering
    \includegraphics[width=\linewidth]{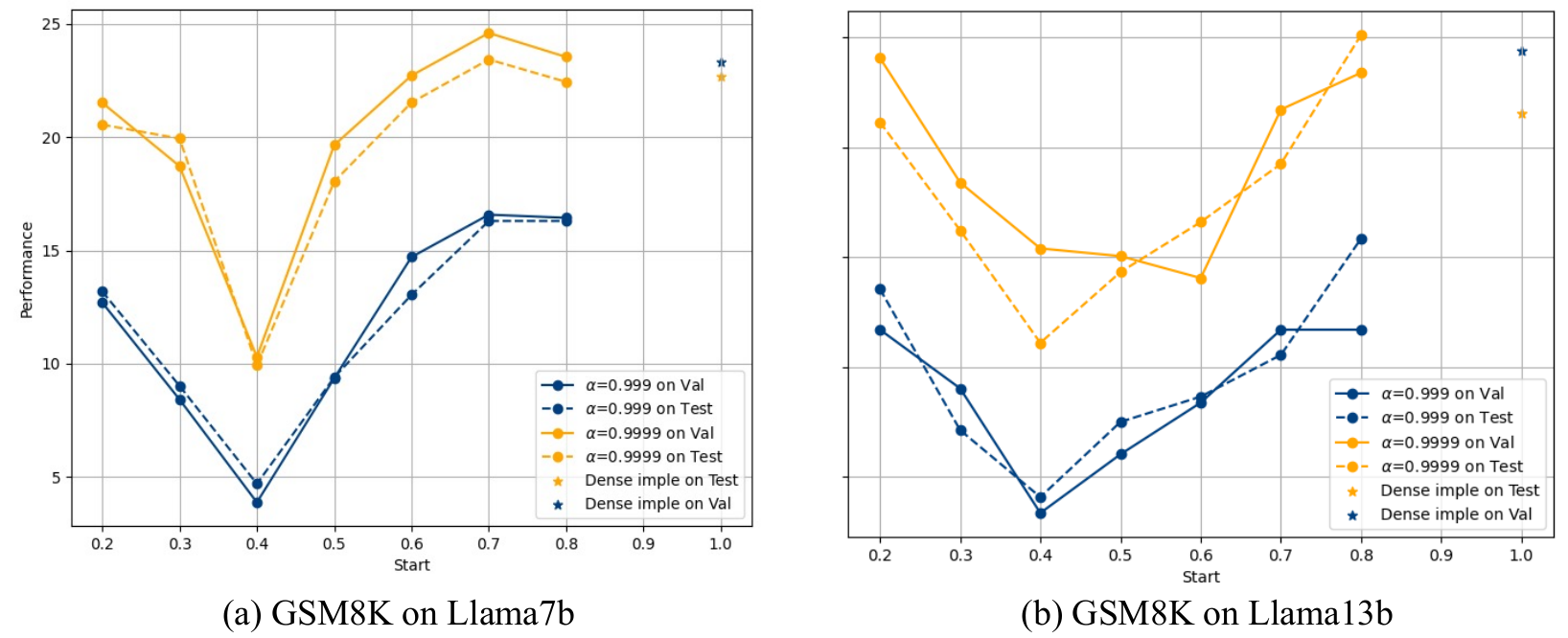}
    \caption{$\alpha$ \textit{vs.} $Start$ accross model scales. The consistent HP sensitivity trends between Validation and Test sets suggest that optimal hyperparameters can be identified on a small validation set and transferred to larger models, saving extensive search efforts.}
    \label{fig:val and test}
\end{figure}

We convert the average skipped layers per token for each task to FLOPs reduction in Table \ref{tab:main_results}. It can be observed that the FLOPs reduction varies for different types of tasks, ranging from $1.07x$ to $1.94x$. This variation is because of different output lengths of each task.  We argue that, \textit{the computation required decreases during the generation process}, and allocating fewer computational resources for later tokens can improve computational efficiency.

Recall that the universality decoder layers, our algorithmic improvements \modelname are fully compatible with traditional model acceleration techniques. This compatibility is noteworthy, especially considering that many earlier instance-wise dynamic depth methods \cite{schuster2022confident} did not support batching and KV cache.
These improvements are applicable to various LLMs and can be easily transferred despite minor architectural and activation function differences.

\begin{figure*}[t!]
    \centering
    \includegraphics[width=\linewidth]{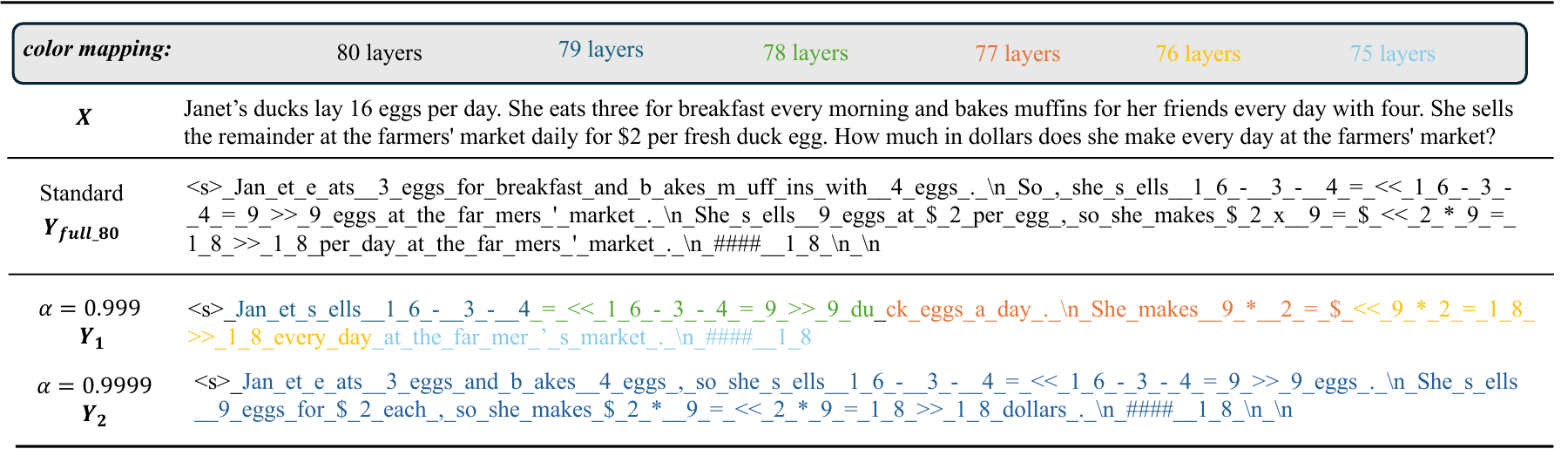}
    \caption{Example output comparison in GSM8K: Hugging Face implementation vs. \modelname outputs under different decay rate $\alpha$ by our design power law decay function $\left\lfloor L \times (\alpha^i) \right\rfloor$. Tokens of different colors correspond to the number of computational layers used for each token $T_i$. 
    When the decay rate is relatively small, the computation decreases rapidly, as seen in the output of $Y_1$. Conversely, for $Y_2$, the computation decreases more slowly.}
    \label{fig: case study}
\end{figure*}

\paragraph{Wall-clock time compared with HuggingFace implementation.}
It's worth mentioning that the actual speedup may slightly vary during generation, given the impossibility of predicting the number of tokens the model will generate in advance \cite{corro2024skipdecode}. Additionally, the introduction of certain computations, such as conditional statements and array creation, also adds to the computational load. Therefore, to accurately assess the practical acceleration benefits of our approach, we integrated \modelname into the lm-harness evaluation framework and conducted end-to-end speed measurements for the entire GSM8K task within the framework, comparing it with the HF implementation. For the LLama 7B and 13B versions, we utilized 2xV100 GPUs with 32GB memory, setting the batch size to 4. As for the 70B model, we employed 8xA100 GPUs with 40GB memory, setting the batch size to 8. All three models were subjected to data parallelism during inference. The results are presented in Table \ref{tab: wall clock}, indicating that our method achieves significant acceleration in real-world scenarios, with this advantage becoming more pronounced as the batch size increases.

\subsection{Factor Study}
We investigate the impact of Flex Layer start position and decay rate on performance and conduct case studies on output samples for different decay rates.

\paragraph{Exploration of decay rate and flex Layer impact.}
Firstly, we examine the effect of decay rate $\alpha$ and Flex Layer $start$ depth on performance on GSM8K, as shown in Figure \ref{fig:val and test}. We draw the following conclusions: (\romannumeral 1) At the same decay rate, it is more appropriate to start the decay from the middle of the model for models of various sizes. This experimental conclusion confirms our hypothesis in Section \ref{sec: core} that LLMs have Core and Flex Layers. The Core Layer generally represents the layers at the beginning and end, while the Flex Layer represents the middle layers, where redundant layers exist and can be skipped to improve inference efficiency. (\romannumeral 2) Additionally, the results presented in the Appendix Table \ref{tab: HP description} for BBH exhibit a similar trend, with optimal hyperparameters differing from those for GSM8K. In this appendix, we provide the detailed numerical results for the BBH benchmark, as shown in Table \ref{tab: bbh em}. Table \ref{tab: HP description} presents the description and search range for each parameter during the hyperparameter tuning process. Table \ref{tab:hyperparam_search_results} demonstrates the successful transfer of optimal parameters, identified using 10\% of the validation set, from smaller models to larger-scale models.

This observation suggests that $\alpha$ and $start$ are task-specific parameters. We hypothesize that this is due to the BBH task having a much shorter output length compared to GSM8K, which necessitates a faster decay rate (i.e., a smaller decay coefficient) for optimal performance.

\paragraph{Example Output: optimizing model capacity allocation by token position.}

Figure \ref{fig: case study} illustrates two example outputs from \modelname for mathematical reasoning responses on the GSM8K dataset, compared to the outputs of the full model (Llama70B with 80 layers) implemented by Huggingface. Black tokens indicate passage through all layers, while other colors represent the number of layers each token has traversed, as computed under different decay coefficients.

\section{Conclusion}
This work introduces a token-position wise layer skipping framework called \modelname which saves 1.5x times operations efficiently while maintaining competitive performance. 
Through analysis of the missing states and input-output flow, we design a training-free algorithm using the power law decay function to decide how many layers to keep per generated token.
Experimental results on mainstream LLMs, demonstrate \modelname significantly improves speed (average 1.5x) compared to HuggingFace implementation on GSM8K and BBH benchmarks, while keeping performance barely no drop. Additionally, \modelname can complement other model acceleration techniques, such as batch processing and KV caching, potentially enhancing inference efficiency. We argue that \modelname establishes a new paradigm for efficient inference alongside existing effective methods.

\section{Limitation}

This work approaches LLM behavior from a global perspective, analyzing per-token perplexity (PPL) dynamics. However, it may overlook the impact of certain special tokens, which could introduce nuances that are not fully captured in the proposed framework.


\section*{Acknowledgments}

This work is supported by the National Science and Technology Major Project (No. 2022ZD0116314), the National Science Foundation of China (No. 62106249).

\bibliography{acl}

\appendix
\newpage



\definecolor{darkgreen}{rgb}{0.0, 0.5, 0.0}
\definecolor{darkred}{rgb}{0.5, 0.0, 0.0}
\begin{table*}[t!]
    \centering
    \resizebox{0.89\textwidth}{!}{
    \begin{tabular}{lcccccc}
        \toprule
        \textbf{Tasks} 
        & \textbf{7B} & \textbf{\modelname-7B} & \textbf{13B} & \textbf{\modelname-13B} & \textbf{70B} & \textbf{\modelname-70B} \\
        \midrule
boolean expressions & 68.40 & 68.40  & 72.80 & 72.80  & 82.00 & 82.00  \\
causal judgement & 50.27 & 50.27  & 54.55 & 54.55  & 62.57 & 62.57 \\
date understanding & 36.80 & \textcolor{darkgreen}{37.60 $\uparrow$ (0.8)} & 51.20 & \textcolor{darkgreen}{51.60 $\uparrow$ (0.4)} & 61.60 & 61.60  \\
disambiguation qa & 53.60 & \textcolor{darkred}{49.20 $\downarrow$ (4.4)} & 32.80 & \textcolor{darkgreen}{36.40 $\uparrow$ (3.6)} & 57.20 & \textcolor{darkgreen}{58.00 $\uparrow$ (0.8)} \\
dyck languages & 8.00 & \textcolor{darkgreen}{19.60 $\uparrow$ (11.6)} & 6.00 & \textcolor{darkred}{7.60 $\downarrow$ (1.6)} & 20.00 & \textcolor{darkgreen}{22.80 $\uparrow$ (2.8)} \\
formal fallacies & 43.60 & 43.60 & 52.40 & 52.40  & 52.00 & \textcolor{darkred}{51.60 $\downarrow$ (0.4)} \\
geometric shapes & 9.20 & \textcolor{darkgreen}{9.60 $\uparrow$ (0.4)} & 31.60 & \textcolor{darkgreen}{34.00 $\uparrow$ (2.4)} & 47.60 & \textcolor{darkred}{46.80 $\downarrow$ (0.8)} \\
hyperbaton & 48.40 & 48.40  & 61.20 & 61.20  & 74.00 & \textcolor{darkred}{72.80 $\downarrow$ (1.2)} \\
logical deduction (5 objects) & 25.60 & \textcolor{darkred}{24.00 $\downarrow$ (1.6)} & 21.20 & \textcolor{darkgreen}{25.60 $\uparrow$ (4.4)} & 35.20 & \textcolor{darkgreen}{36.00 $\uparrow$ (0.8)} \\
logical deduction (7 objects) & 14.80 & \textcolor{darkgreen}{15.20 $\uparrow$ (0.4)} & 18.40 & \textcolor{darkgreen}{24.00 $\uparrow$ (5.6)} & 42.00 & \textcolor{darkred}{41.20 $\downarrow$ (0.8)} \\
logical deduction (2 objects) & 32.40 & \textcolor{darkgreen}{36.00 $\uparrow$ (3.6)} & 39.60 & \textcolor{darkgreen}{41.60 $\uparrow$ (2)} & 59.20 & \textcolor{darkgreen}{63.60 $\uparrow$ (4.4)} \\
movie recommendation & 38.40 & \textcolor{darkgreen}{41.60 $\uparrow$ (3.2)} & 74.40 & \textcolor{darkgreen}{76.00 $\uparrow$ (1.6)} & 92.00 & \textcolor{darkgreen}{92.80 $\uparrow$ (0.8)} \\
multistep arithmetic two & 0.40 & \textcolor{darkgreen}{0.80 $\uparrow$ (0.4)} & 1.20 & 1.20& 1.20 & \textcolor{darkgreen}{1.60 $\uparrow$ (0.4)} \\
navigate & 41.60 & 41.60 & 59.20 & 59.20 & 59.60 & \textcolor{darkred}{58.80 $\downarrow$ (0.8)} \\
object counting & 35.60 & \textcolor{darkred}{28.80 $\downarrow$ (6.8)} & 50.00 & \textcolor{darkred}{48.40 $\downarrow$ (1.6)} & 51.60 & \textcolor{darkgreen}{53.20 $\uparrow$ (1.6)} \\
penguins in a table & 24.66 & \textcolor{darkgreen}{27.40 $\uparrow$ (2.74)} & 29.45 & \textcolor{darkgreen}{30.14 $\uparrow$ (0.69)} & 39.04 & \textcolor{darkgreen}{41.10 $\uparrow$ (2.06)} \\
reasoning about colored objects & 21.60 & \textcolor{darkred}{21.20 $\downarrow$ (0.4)} & 27.20 & \textcolor{darkred}{26.80 $\downarrow$ (0.4)} & 45.60 & \textcolor{darkgreen}{49.20 $\uparrow$ (3.6)} \\
ruin names & 26.40 & \textcolor{darkgreen}{27.60 $\uparrow$ (1.2)} & 40.00 & \textcolor{darkgreen}{43.60 $\uparrow$ (3.6)} & 82.80 & \textcolor{darkgreen}{84.80 $\uparrow$ (2)} \\
salient translation error detection & 23.60 & \textcolor{darkred}{22.00 $\downarrow$ (1.6)} & 35.20 & \textcolor{darkred}{34.00 $\downarrow$ (1.2)} & 50.40 & 50.40\\
snarks & 46.63 & \textcolor{darkgreen}{48.88 $\uparrow$ (2.25)} & 52.81 & \textcolor{darkgreen}{53.37 $\uparrow$ (0.56)} & 80.34 & \textcolor{darkgreen}{80.90 $\uparrow$ (0.56)} \\
word sorting & 12.00 & \textcolor{darkgreen}{20.00 $\uparrow$ (10)} & 16.00 & \textcolor{darkgreen}{31.60 $\uparrow$ (15.6)} & 32.40 & \textcolor{darkred}{27.20 $\downarrow$ (5.2)} \\
sports understanding & 66.40 & 66.40 & 64.80 & 64.80  & 80.40 &80.40 \\
temporal sequences & 7.60 & \textcolor{darkgreen}{10.00 $\uparrow$ (2.4)} & 18.80 & \textcolor{darkred}{12.00 $\downarrow$ (6.8)} & 62.40 & \textcolor{darkgreen}{66.80 $\uparrow$ (4.4)} \\
tracking shuffled objects (5 objects) & 15.60 & \textcolor{darkgreen}{16.8 $\uparrow$ (1.2)} & 17.20 & \textcolor{darkred}{16.40 $\downarrow$ (0.8)} & 16.40 & \textcolor{darkred}{15.60 $\downarrow$ (0.8)} \\
tracking shuffled objects (7 objects) & 16.80 & \textcolor{darkred}{11.60 $\downarrow$ (5.2)} & 12.80 & \textcolor{darkgreen}{14.80 $\uparrow$ (2)} & 14.00 & \textcolor{darkgreen}{15.60 $\uparrow$ (1.6)} \\
tracking shuffled objects (2 objects) & 33.20 & 33.20 & 32.00 & \textcolor{darkgreen}{33.20 $\uparrow$ (1.2)} & 30.40 & \textcolor{darkred}{29.20 $\downarrow$ (1.2)} \\
web of lies & 48.80 & 48.80 & 52.00 & 52.00 & 48.80 & 48.80  \\
\midrule
 Avg. 27 Tasks & 31.22 & \textcolor{darkgreen}{31.88 $\uparrow$ (0.66)} & 37.77 & \textcolor{darkgreen}{39.07 $\uparrow$ (1.3)} & 50.90 & \textcolor{darkgreen}{51.42 $\uparrow$ (0.52)}\\

        \bottomrule
    \end{tabular}}
    \caption{Detailed Performance comparison for BBH 27 tasks between HuggingFace Implementation and \modelname}
    \label{tab: bbh em}
\end{table*}

\begin{table*}[t!]
    \centering
    \resizebox{0.8\textwidth}{!}{
    \begin{tabular}{lccc}
        \toprule
        \textbf{$T_i$ Drop Layer set} & \textbf{HP} & \textbf{Grid Search Space} & \textbf{Description} \\
        \midrule
       \multirow{2}{*}{\scriptsize $\{ l \mid $start$< l < $start$ + L - \left\lfloor L \times (\alpha^i) \right\rfloor \}$} 
        &  $start$ & \scriptsize [0.2, 0.3, 0.4, 0.5, 0.6, 0.7, 0.8] & Drop layer start position \\
        &  $\alpha$ & \scriptsize [0.999, 0.9999] & Decay rate \\
        \bottomrule
    \end{tabular}}
     \caption{Details of the computation layer ID set used for generating token $T_i$ at time step $i$, along with the search range and descriptions for the formula's hyperparameters.}
     \label{tab: HP description}
\end{table*}

\begin{table*}[t!]
    \centering
    \resizebox{0.89\textwidth}{!}{
    \begin{tabular}{lcccccccccccccc}
        \toprule
        \multirow{3}{*}{\bf{$\alpha$}} & \multicolumn{14}{c}{\bf{$start$}} \\
        \cmidrule(lr){2-15}
        & \multicolumn{2}{c}{\textbf{0.2}} & \multicolumn{2}{c}{\textbf{0.3}} & \multicolumn{2}{c}{\textbf{0.4}} & \multicolumn{2}{c}{\textbf{0.5}} & \multicolumn{2}{c}{\textbf{0.6}} & \multicolumn{2}{c}{\textbf{0.7}} & \multicolumn{2}{c}{\textbf{0.8}} \\
        \cmidrule(lr){2-3} \cmidrule(lr){4-5} \cmidrule(lr){6-7} \cmidrule(lr){8-9} \cmidrule(lr){10-11} \cmidrule(lr){12-13} \cmidrule(lr){14-15}
        & \textbf{Val} & \textbf{Test} & \textbf{Val} & \textbf{Test} & \textbf{Val} & \textbf{Test} & \textbf{Val} & \textbf{Test} & \textbf{Val} & \textbf{Test} & \textbf{Val} & \textbf{Test} & \textbf{Val} & \textbf{Test} \\
        \midrule
        \bf \textit{Llama 7b} & \multicolumn{14}{c}{Dense Implementation of Validation and Test Results: 11.76, 10.61} \\
        {0.999}   &6.68	&7.42 &5.61&4.85	&3.34&3.63&	4.41&5.00&	5.35&5.46	&6.68&6.22&	6.68&8.34 \\
                               
        {0.9999} &\cellcolor{blue!20}11.63&\cellcolor{gray!20}10.46&	9.36&8.49&	8.16&6.44&	8.02&7.66&	7.62&8.42&	10.69	&9.70&\cellcolor{blue!20}11.36 &\cellcolor{gray!20}12.05 \\
                          
        \bf \textit{Llama 13b} & \multicolumn{14}{c}{Dense Implementation of Validation and Test Results: 23.32, 22.67} \\
        {0.999}  &12.70	&13.19 &8.42	& 9.02&3.88&4.70 &	9.36& 9.40&	14.71&13.04 &	16.58&16.30 &	16.44&16.30 \\
                               
        {0.9999} &\cellcolor{blue!20} 21.52&\cellcolor{gray!20}20.55&18.72&19.94&	10.30	&9.33&19.65	&18.04&22.72	&21.53&\cellcolor{blue!20}24.60&\cellcolor{gray!20}23.43&	23.54&22.44\\
        \bottomrule
    \end{tabular}}
   	\caption{Hyperparameter Search Results: Performance for Different $start$ and $\alpha$ Values, Including Validation (10\% of train set) and Test Results. The results indicate that \textbf{minor searches in validation parameters are effective on test performance, showing consistent sensitivity trends}. Due to space constraints, only the GSM8k results are shown; similar results apply to the BBH task. We highlighted the best parameter combinations in Table 1, noting that a broader search could yield even better results.}

    \label{tab:hyperparam_search_results}
\end{table*}

\end{document}